\documentclass{article}

\usepackage{arxiv}

\usepackage[utf8]{inputenc} 
\usepackage[T1]{fontenc}    
\usepackage{hyperref}       
\usepackage{url}            
\usepackage{booktabs}      
\usepackage{amsfonts}       
\usepackage{nicefrac}       
\usepackage{lipsum}
\usepackage{natbib}
\usepackage{doi}
\usepackage{microtype}
\usepackage{tabularx}
\usepackage{graphicx}
\usepackage{amsmath}
\usepackage{amssymb}
\usepackage{algorithm}
\usepackage{algorithmic}
\usepackage{lineno}
\usepackage{enumitem}
\usepackage{multirow}
\usepackage{booktabs}
\usepackage{xcolor}
\usepackage{pifont}
\usepackage[section]{placeins}

\title{Bench2Robust: Scenario-Controlled Recovery for Tool-Using Agents}

\date{}

\author{
\bfseries
Chaoran Chen \quad
Vy Nguyen \quad
Ziji Zhang \quad
Abhinav Gullapalli \quad
Ziyi Wang
\\[2pt]
\bfseries
Yuxuan Lu \quad
Dakuo Wang \quad
Jing Huang \quad
Zhou Yu \quad
Jin Lai
\\[2pt]
\normalfont Amazon
}

\renewcommand{\headeright}{}
\renewcommand{\undertitle}{}

\newcommand{\bench}{\textsc{Bench2Robust}}
\newcommand{\taubench}{$\tau^2$-bench}

\begin{document}

\maketitle

\begin{abstract}
Tool-using LLM agents are typically developed and post-trained in controlled environments with reliable tools, but deployment exposes them to tool failures, API and schema drift, and third-party services outside the developer's control. We introduce \bench{}, a framework for scenario-controlled solvability that augments existing benchmarks with episodes requiring retrying the original path, switching to a fallback-equivalent tool, or recognizing that no viable path remains. By controlling recoverability, \bench{} separates recovery-policy quality from randomness in whether failures persist or resolve. Across 7 models from 4 families and 10 task/domain slices spanning customer-service and function-calling benchmarks, 69 of 70 model-subset pairs degrade under tool failures, by up to 46.7 percentage points. We propose Bayesian Tool Memory (BTM) for structured runtime recovery context and use reinforcement learning (RL) for learned recovery behavior. On held-out Retail tasks using the benchmark's original tool set, BTM improves robustness by 16.8 percentage points without retraining, while RL retains a 6.3-point gain without inference-time BTM. Combining RL with BTM yields a 20.7-point improvement over the base model while preserving clean performance. These results show that robust tool use benefits from combining structured runtime recovery context with learned recovery behavior.
\end{abstract}

\section{Introduction}
\label{sec:intro}

\begin{figure}[t]
    \centering
    \includegraphics[width=\textwidth]{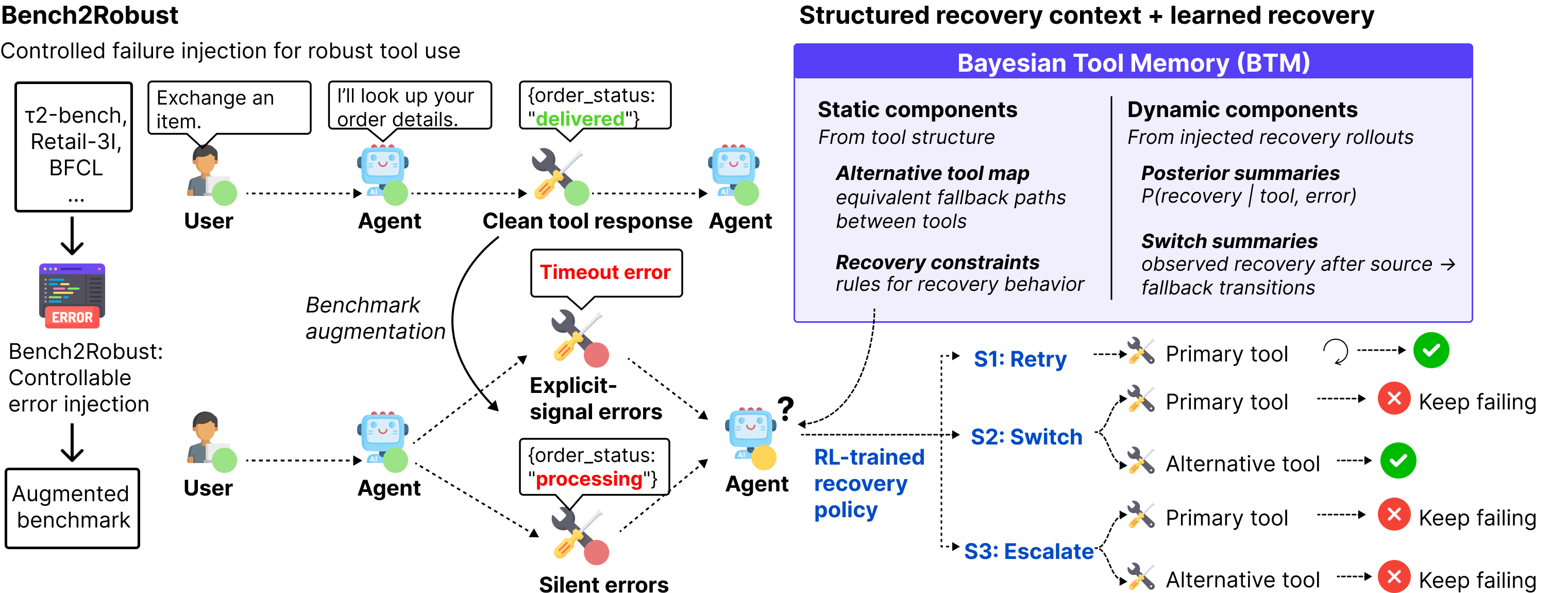}
    \caption{
    \textbf{Overview of \bench{} and our recovery approach.}
    \bench{} augments existing tool-use benchmarks with controllable tool-response failures, including explicit-signal errors and silent corruptions. Bayesian Tool Memory (BTM) combines static recovery structure---an alternative-tool map and recovery constraints---with dynamic recovery statistics estimated from injected recovery rollouts.
    }
    \label{fig:overview}
\end{figure}

Large language models (LLMs) augmented with tools can complete complex multi-turn tasks by retrieving information and acting through external systems~\citep{schick2023toolformer,qin2024toolllm,patil2024gorilla}. Most tool-use benchmarks, however, assume that tool calls return timely, structurally valid, correct, and current responses~\citep{yao2025taubench,li2023apibank}. Real tools violate these assumptions: requests time out or hit rate limits, credentials and schemas become invalid, and apparently valid responses may contain stale, partial, or incorrect data. Moreover, tool failures vary in both persistence and observability: some resolve on retry, some persist, and others provide no explicit failure signal. Robust agents must therefore reason about what remains recoverable under changing tool conditions and choose an appropriate response: \textsc{Retry} the current path, \textsc{Switch} to an alternative, or \textsc{Abstain} and escalate once viable paths are exhausted (Figure~\ref{fig:overview}).

A central challenge is that conventional stochastic failure injection does not specify what recovery behavior an episode actually requires. If a transient failure happens to disappear on the next call, indiscriminate retrying may appear robust. Conversely, a reasonable recovery action may still receive low reward if the sampled failure persists. As a result, observed task success can conflate recovery-policy quality with stochastic variation in the environment, making it difficult to ask a more fundamental question: \emph{does the agent choose a recovery strategy that matches what remains solvable?}

We introduce \bench{}, a benchmark-agnostic framework built around \textbf{scenario-controlled solvability}. Alongside stochastic tool-response corruption, \bench{} assigns each episode to one of three controlled recoverability scenarios: in S1, the original path remains viable and retrying can recover; in S2, the primary path is blocked episode-wide and completion requires a fallback-equivalent tool; and in S3, all viable paths are blocked, making escalation the appropriate response. By fixing the recoverability scenario at the environment level, \bench{} separates whether a task remains solvable from how the agent responds, establishing a controlled setting for both evaluation and recovery-oriented training. All three scenarios are used during training. Our held-out task-success evaluation focuses on solvable settings because existing benchmark evaluators do not reward correct stopping on intentionally unsolvable S3 episodes, and we avoid adding an explicit abstention reward that could incentivize premature stopping.

We use \bench{} to study two complementary approaches to recovery. \emph{Bayesian Tool Memory} (BTM) provides structured runtime context combining fallback relations, recovery constraints, and empirical recoverability statistics, while reinforcement learning trains recovery behavior across controlled scenarios. Our primary policy is Qwen3-4B, with a Qwen3-8B replication (Appendix~\ref{app:8b_results}); BTM is used during RL exploration and optionally at inference.

We evaluate robustness across \taubench{}~\citep{barres2025tau2bench}, the Retail-3I and Airline-3I that extend the corresponding \taubench{} domains~\citep{wang-etal-2026-trajectory2task}, and BFCL multi-turn~\citep{patil2025bfcl}, covering seven models from four model families. Recovery interventions are trained on Combined Retail (1,339 tasks from \taubench{} Retail and Retail-3I) and evaluated on a 402-task held-out split, with Airline-3I, Telecom, and BFCL used for cross-benchmark evaluation.

Our contributions are threefold. First, we formulate tool-failure robustness as recovery under \emph{controlled solvability}, explicitly separating episodes where retrying remains viable, switching is required, or no viable path remains. Second, we instantiate this framework across multiple tool-use benchmark families and characterize a broad robustness gap across models, domains, and failure regimes. Third, we study how structured runtime recovery context and RL training
contribute separately and jointly to robustness on held-out tasks.

Our main findings are:
\begin{itemize}[leftmargin=1.2em, itemsep=0.25em, topsep=0.25em]
    \item \textbf{Tool failures expose a broad robustness gap.} Across 7 models, 4 families, and 10 task/domain slices, 69 of 70 model-subset pairs degrade under injection, by up to 46.7pp.

    \item \textbf{Structured recovery context provides large zero-training gains.} Online-updated BTM improves the 4B base model by +16.8pp without alternatives and +11.7pp with alternatives; ablations separate complementary contributions from static recovery structure and online adaptation.

    \item \textbf{RL produces recovery gains that persist without inference-time BTM.} Removing BTM at inference, the trained policy still improves robustness by +6.3/+6.9pp over the base model.

    \item \textbf{Runtime context and learned behavior contribute differently across error types.} BTM provides the largest gains on several explicit service/response errors, while the additional RL-associated gain is larger on auth/schema errors and factual corruption.
\end{itemize}

\section{The \bench{} Framework}
\label{sec:method_framework}

\bench{} is a benchmark-agnostic injection framework that interposes between agent and tool execution, using a domain registry of task-relevant fallback relations and viable paths. Benchmarks following the \texttt{tool\_call(name, args) $\to$ response} pattern can be integrated via a lightweight adapter; we support \taubench{} (including Retail-3I/Airline-3I) and BFCL. We model 9 tool-response failures: 6 explicit-signal failures that suppress execution (timeout, rate limit, server error, auth error, malformed response, schema drift) and 3 silent corruptions that alter only the returned observation (partial data, stale values, factual errors). For calls not subject to scenario blocking, each tool call samples $\nu_t \sim \text{Cat}(\boldsymbol{\theta}_j)$ with $P(\text{clean})=0.60$; the POMDP formalization appears in Appendix~\ref{app:injection_algorithm}, with the full noise configuration in Appendix~\ref{app:injection}.

\subsection{Scenario-Controlled Solvability}
\label{sec:scenarios}

The central design choice is to assign each training episode an explicit recoverability scenario. In \textbf{S1} (original path viable), no task-relevant path is permanently blocked, so the original path remains viable despite stochastic tool failures. In \textbf{S2} (switch needed), the task has multiple fallback-equivalent solving paths. We randomly select one path to block; task completion therefore requires switching to a remaining fallback-equivalent path. In \textbf{S3} (impossible), all task-relevant viable paths are blocked for the entire episode, so continued tool use cannot complete the task and escalation is the appropriate prescribed behavior. Scenario blocking is deterministic and takes precedence over stochastic failure injection: calls to blocked tools always return a blocked response, whereas calls to unblocked tools remain subject to the per-call stochastic noise distribution with $P(\text{clean})=0.60$. Blocking therefore operates outside the stochastic injection budget ($B \leq 5$, $K_{\max}=2$ consecutive stochastic failures on the same tool), separating episode-level path unavailability from an unlucky sequence of transient failures. Algorithm~\ref{alg:bench2robust} formalizes this procedure. In the current evaluation, S1 and S2 can be scored through task completion, whereas S3 is used during training because the underlying benchmark evaluators do not assign positive completion credit to unsolvable tasks.

\begin{algorithm}[t]
\small
\caption{\bench{} Scenario-Controlled Failure Injection}
\label{alg:bench2robust}
\begin{algorithmic}[1]
\REQUIRE Task $\tau$, tool set $\mathcal{T}$, task-relevant fallback classes $\mathcal{E}_{\tau}$, viable tool paths $\mathcal{V}_{\tau}$, scenario $s \in \{\mathrm{S1,S2,S3}\}$, per-tool noise distributions $\{\boldsymbol{\theta}_j\}_{j\in\mathcal{T}}$, execution-suppressing/silent noise partition $\mathcal{N}_{\mathrm{exec}}, \mathcal{N}_{\mathrm{silent}}$, budget $B$, consecutive limit $K_{\max}$
\STATE $\mathcal{P} \leftarrow \emptyset$
\IF{$s=\mathrm{S2}$}
    \STATE Select a task-relevant equivalence class $E \in \mathcal{E}_{\tau}$
    \STATE Randomly choose one side $j \in E$ and set $\mathcal{P}\leftarrow\{j\}$
\ELSIF{$s=\mathrm{S3}$}
    \STATE $\mathcal{P}\leftarrow \mathcal{V}_{\tau}$
\ENDIF
\FOR{each tool call $(j,a)$ in the episode}
    \IF{$j\in\mathcal{P}$}
        \STATE Return episode-persistent blocked response \COMMENT{$(j,a)$ not executed; state unchanged}
    \ELSIF{stochastic injection budget remains and consecutive failures for $j$ $< K_{\max}$}
        \STATE Sample $\nu \sim \mathrm{Cat}(\boldsymbol{\theta}_j)$
        \IF{$\nu \in \mathcal{N}_{\mathrm{exec}}$}
            \STATE Return failure message for $\nu$ and update stochastic injection counters \COMMENT{$(j,a)$ not executed; state unchanged}
        \ELSIF{$\nu \in \mathcal{N}_{\mathrm{silent}}$}
            \STATE $o^{\star}\leftarrow Z_{\mathrm{clean}}(s_t,(j,a))$ \COMMENT{$(j,a)$ executes normally}
            \STATE Return $C_{\nu}(o^{\star})$ and update stochastic injection counters
        \ENDIF
    \ENDIF
    \STATE $o^{\star}\leftarrow Z_{\mathrm{clean}}(s_t,(j,a))$ \COMMENT{$(j,a)$ executes normally}
    \STATE Return clean response $o^{\star}$
\ENDFOR
\end{algorithmic}
\end{algorithm}

\subsection{Alternative Tool Paths}
\label{sec:alt_tools}

For Retail we add 5 alternative tools (21 total), each providing a different query pattern to the same data (e.g., \texttt{search\_product\_by\_name} as alternative to \texttt{get\_product\_details}). For BFCL evaluation, we identify natural equivalence classes within each domain and apply the same S1/S2/S3 blocking. Full specifications in Appendix~\ref{app:alt_tools} (Table~\ref{tab:alt_tools_retail}).

\section{Strategy-Aware Training}
\label{sec:training}

\subsection{Bayesian Tool Memory (BTM)}
\label{sec:belief_provider}

Random rollouts from a base model rarely contain enough successful recovery behavior to provide useful RL signal, so we supply \emph{Bayesian Tool Memory} (BTM) as structured recovery context during training and optionally at inference. BTM combines (1) Beta-posterior recoverability statistics estimated from training-split rollouts, (2) domain-specific fallback maps with granularity annotations, and (3) recovery constraints for retrying, verification, and escalation.

For each (tool, error type) pair we estimate a Beta-posterior episode-level recovery probability from base-model rollouts on the Retail training split (held-out never used), together with analogous switch-transition beliefs for (source, target) tool pairs; full posterior definitions, sample counts, and the prompt template are in Appendix~\ref{app:btm_prompt}. These quantities are empirical episode-level recoverability estimates, not causal probabilities that a particular retry or switch action will succeed. During Retail RL training, BTM serves as exploration context that makes retry, switching, verification, and escalation behaviors more likely to appear in rollouts, creating trajectories the optimizer can assign credit to; at inference, the same context supplies domain-specific recovery information.

At inference, dynamic beliefs can either be held \emph{fixed} at their training-split estimates or updated \emph{online} across evaluation episodes. In the online setting, after each completed episode, the Beta counts for the relevant tool--error observations are updated according to whether the trajectory ultimately recovered successfully; the resulting posterior is then used as context for subsequent episodes. Each evaluation run starts from the same training-split posterior initialization, with online counts reset between runs. Episodes are processed in fixed order, and updates use the benchmark outcome observed after each completed trajectory. Policy parameters are never updated during evaluation. Static BTM components are fixed in both settings.

\subsection{Reward Design}
\label{sec:reward}

\begin{equation}
R(\tau) = \begin{cases}
1.0 \cdot \text{eff}(\tau) \cdot \text{rep}(\tau) & \text{if task completed} \\
0.3 \cdot \frac{\text{matched actions}}{\text{total required}} \cdot \text{rep}(\tau) & \text{otherwise}
\end{cases}
\label{eq:reward}
\end{equation}
where $\text{eff}(\tau) = \max(0.3,\; 1.0 - 0.02 \cdot \max(0, |\tau| - 12))$ penalizes long episodes and $\text{rep}(\tau) \in \{0, 0.5, 1.0\}$ penalizes repetition ($\geq$4 identical calls $\to$ 0). The partial-credit term provides dense signal: without it, 38\% of training tasks have zero reward variance. Importantly, the reward contains no positive term for correct abstention: S3 episodes cannot pass the benchmark task evaluator. Premature escalation on solvable S1/S2 episodes forfeits completion reward, while the positive prescription to stop after exhausting paths comes from the recovery constraints and exposure to S3 during the curriculum. Accordingly, we treat S3 as a controlled training condition rather than evidence that abstention is independently reward-learned.

\subsection{Curriculum and DAPO Training}
\label{sec:curriculum}

Five phases progressively introduce harder scenarios, from S1-heavy to a mixed S1/S2/S3 curriculum, with phase advancement gated on diagnostic thresholds (e.g., an S3-conditional escalation rate, not a correctness measure) rather than a claim of mastery. The complete phase schedule and advancement diagnostics, including a training-run measurement of escalation rate before and after S3 exposure, are given in Appendix~\ref{app:config}.

We use DAPO~\citep{dapo} with a KL constraint to prevent policy collapse---without it, training collapses within 25--30 iterations as repetition rate exceeds 80\%---together with asymmetric clipping, dynamic group filtering, and group-relative advantages over a hybrid clean/injected rollout batch; full hyperparameters are in Appendix~\ref{app:config}.

\section{Experimental Setup}
\label{sec:evaluation_protocol}

\paragraph{Training and evaluation environments.} Interventions are trained and evaluated on \textbf{Combined Retail}: 1,339 multi-turn customer-service tasks composed of \taubench{} retail (114 tasks)~\citep{barres2025tau2bench} and Retail-3I (1,225 tasks covering general/ambiguous/changing intents)~\citep{wang-etal-2026-trajectory2task}, split into 937 training and 402 held-out tasks. The broader robustness-gap evaluation covers \taubench{} retail/airline/telecom, Retail-3I, Airline-3I, and BFCL multi-turn~\citep{patil2025bfcl}. Cross-benchmark intervention evaluations use Airline-3I (584 tasks), \taubench{} Telecom (114 tasks), and BFCL multi-turn (200 tasks); target-domain BTM, when enabled, is inference-time context and does not update the Retail-trained RL policy.

\paragraph{Evaluation Modes.} (1)~\textbf{Failure-free (clean):} uncertainty injection is disabled; tool calls return the benchmark environment's original responses without artificial delay, failure, or corruption---i.e., timely, structurally valid, correct, and current with respect to the underlying simulator state. (2) \textbf{Inject w/o alt:} injection enabled with the benchmark's standard tool set; scenario control uses only fallback-equivalent paths already present in that tool set. (3)~\textbf{Inject w/ alt:} injection enabled with augmented tool set and scenario-based blocking.

\paragraph{Models.}
Qwen3-4B-Thinking-2507~\citep{yang2025qwen3technicalreport} is our primary training target; we additionally train Qwen3-8B as a replication. Qwen3-8B/32B/235B~\citep{yang2025qwen3technicalreport}, DeepSeek-V3~\citep{deepseekai2025deepseekv3technicalreport}, GLM-4.7~\citep{5team2025glm45agenticreasoningcoding}, and MiniMax-M2.5~\citep{minimax2026minimaxm2seriesminiactivations} serve as reference models for the robustness-gap evaluation, spanning four model families and scales from 4B to 235B parameters. Qwen3-235B serves as the user simulator throughout.

\section{Results}
\label{sec:experiments}

We first characterize the robustness gap, then evaluate the contributions of runtime recovery context and RL, analyze heterogeneity across failure types, and decompose the intervention.

\begin{table}[t]
\centering
\scriptsize
\setlength{\tabcolsep}{3pt}
\renewcommand{\arraystretch}{0.92}
\begin{tabular}{l ccc c ccc c ccc c c}
\toprule
& \multicolumn{3}{c}{\textbf{$\tau^2$-bench}} & & \multicolumn{3}{c}{\textbf{Retail-3I}} & & \multicolumn{3}{c}{\textbf{Airline-3I}} & & \textbf{BFCL} \\
\cmidrule(lr){2-4} \cmidrule(lr){6-8} \cmidrule(lr){10-12} \cmidrule(lr){14-14}
\textbf{Model} & retail & airline & telecom & & general & ambig. & changing & & general & ambig. & changing & & multi-turn \\
& (114) & (50) & (114) & & (473) & (473) & (279) & & (254) & (254) & (76) & & (200) \\
\midrule
\multicolumn{14}{l}{\emph{Clean pass rate (\%)}} \\
Qwen3-4B-Thinking-2507 & 56.1 & 62.0 & 21.1 & & 71.9 & 62.0 & 63.4 & & 65.7 & 61.0 & 39.5 & & 31.5 \\
Qwen3-8B & 41.2 & 32.0 & 25.4 & & 67.0 & 55.8 & 54.5 & & 50.0 & 44.5 & 21.1 & & 26.0 \\
Qwen3-32B & 43.0 & 30.0 & 21.1 & & 70.2 & 58.1 & 53.8 & & 50.0 & 51.6 & 22.4 & & 39.5 \\
Qwen3-235B & 57.9 & 46.0 & 36.8 & & 72.7 & 61.1 & 58.4 & & 67.7 & 65.8 & 39.5 & & 59.5 \\
DeepSeek-V3 & 57.0 & 50.0 & 38.6 & & 74.0 & 62.4 & 62.0 & & 68.1 & 68.5 & 44.7 & & 50.0 \\
GLM-4.7 & 60.5 & 64.0 & 89.5 & & 72.7 & 60.0 & 61.7 & & 63.0 & 66.9 & 31.6 & & 77.5 \\
MiniMax-M2.5 & 64.9 & 68.0 & 91.2 & & 71.0 & 61.1 & 61.7 & & 67.3 & 68.9 & 42.1 & & 81.5 \\
\midrule
\multicolumn{14}{l}{\emph{Inject pass rate (\%)}} \\
Qwen3-4B-Thinking-2507 & 15.8 & 52.0 & 19.3 & & 25.2 & 19.7 & 28.7 & & 37.0 & 31.1 & 7.9 & & 13.5 \\
Qwen3-8B & 17.5 & 38.0 & 21.1 & & 41.9 & 29.8 & 31.2 & & 46.1 & 37.8 & 11.8 & & 14.5 \\
Qwen3-32B & 21.1 & 22.0 & 16.7 & & 49.1 & 37.8 & 41.6 & & 39.4 & 41.3 & 11.8 & & 19.0 \\
Qwen3-235B & 42.1 & 40.0 & 25.4 & & 58.8 & 44.6 & 43.0 & & 55.1 & 55.1 & 29.0 & & 26.0 \\
DeepSeek-V3 & 37.7 & 36.0 & 29.8 & & 54.1 & 43.6 & 43.7 & & 54.7 & 57.9 & 39.5 & & 29.0 \\
GLM-4.7 & 29.8 & 52.0 & 68.4 & & 43.8 & 39.8 & 39.1 & & 43.3 & 45.7 & 19.7 & & 38.5 \\
MiniMax-M2.5 & 56.1 & 52.0 & 71.1 & & 61.3 & 46.7 & 52.3 & & 54.3 & 61.0 & 34.2 & & 54.5 \\
\midrule
\multicolumn{14}{l}{\emph{$\Delta$ (pp)}} \\
Qwen3-4B-Thinking-2507 & $-$40.4 & $-$10.0 & $-$1.8 & & $-$46.7 & $-$42.3 & $-$34.8 & & $-$28.7 & $-$29.9 & $-$31.6 & & $-$18.0 \\
Qwen3-8B & $-$23.7 & +6.0$^\dagger$ & $-$4.4 & & $-$25.2 & $-$26.0 & $-$23.3 & & $-$3.9 & $-$6.7 & $-$9.2 & & $-$11.5 \\
Qwen3-32B & $-$21.9 & $-$8.0 & $-$4.4 & & $-$21.1 & $-$20.3 & $-$12.2 & & $-$10.6 & $-$10.2 & $-$10.5 & & $-$20.5 \\
Qwen3-235B & $-$15.8 & $-$6.0 & $-$11.4 & & $-$14.0 & $-$16.5 & $-$15.4 & & $-$12.6 & $-$10.6 & $-$10.5 & & $-$33.5 \\
DeepSeek-V3 & $-$19.3 & $-$14.0 & $-$8.8 & & $-$19.9 & $-$18.8 & $-$18.3 & & $-$13.4 & $-$10.6 & $-$5.3 & & $-$21.0 \\
GLM-4.7 & $-$30.7 & $-$12.0 & $-$21.1 & & $-$29.0 & $-$20.3 & $-$22.6 & & $-$19.7 & $-$21.3 & $-$11.8 & & $-$39.0 \\
MiniMax-M2.5 & $-$8.8 & $-$16.0 & $-$20.2 & & $-$9.7 & $-$14.4 & $-$9.3 & & $-$13.0 & $-$7.9 & $-$7.9 & & $-$27.0 \\
\bottomrule
\end{tabular}
\caption{Robustness gap across four task suites (\% pass rate, temp=0, inject w/o alt). The first three suites use $\tau^2$-bench-style customer-service infrastructure; BFCL is an independent multi-turn function-calling benchmark. $^\dagger$Small positive delta on $\tau^2$-bench Airline ($n$=50) is within noise.}
\label{tab:cross_benchmark}
\end{table}
\renewcommand{\arraystretch}{1}

\subsection{The Robustness Gap Is Broad}
\label{sec:cross_benchmark}

We evaluate 7 models from 4 families on 10 task/domain slices under injection (Table~\ref{tab:cross_benchmark}). Of 70 (model, subset) pairs, 69 degrade ($-$1.8 to $-$46.7pp). The gap appears across customer service, airline booking, and function-calling domains. Larger models sometimes lose fewer absolute points, but scaling does not eliminate brittleness: Qwen3-235B still loses 15.8pp on $\tau^2$-retail, while models with 89--91\% clean Telecom performance (GLM-4.7, MiniMax-M2.5) lose 20--21pp. BFCL also shows large drops ($-$11.5 to $-$39pp). Because task structure and model variants differ across rows, we interpret these results as evidence of broad failure sensitivity rather than as a controlled scaling law; the smallest subsets ($n{=}50$--$114$) carry a $\sim$5--7pp noise floor, so the sole non-degrading cell and several smaller deltas ($-$1.8 to $-$5.3pp) are not reliably distinguishable from zero.

\subsection{Within-Benchmark Effectiveness}
\label{sec:main_results}

We next evaluate how much robustness comes from structured runtime recovery context and how much additional benefit is associated with RL training. Table~\ref{tab:main_results} reports held-out results (402 tasks not seen during RL training or belief computation) for both the primary 4B target and the Qwen3-8B replication, as point estimates; full mean$\pm$std over 5 evaluation seeds for the 4B rows appears in Appendix Table~\ref{tab:main_results_std}. RL models are trained with BTM as an exploration prior; we evaluate both with and without BTM at inference time to measure internalization.

\begin{table}[!htbp]
\centering
\small
\renewcommand{\arraystretch}{0.94}
\begin{tabular}{l cc ccc}
\toprule
\textbf{Model} & \textbf{Train BTM} & \textbf{Infer BTM} & \textbf{Clean} & \textbf{Inj w/o Alt} & \textbf{Inj w/ Alt} \\
\midrule
4B & \ding{55} & \ding{55} & 64.3 & 20.1 & 31.9 \\
4B & \ding{55} & \ding{51} & 65.3 & 36.9 & 43.6 \\
4B + RL & \ding{51} & \ding{55} & 64.5 & 26.4 & 38.8 \\
4B + RL & \ding{51} & \ding{51} & 63.9 & \textbf{40.8} & \textbf{45.5} \\
\midrule
8B & \ding{55} & \ding{55} & 58.3 & 33.3 & 39.4 \\
8B & \ding{55} & \ding{51} & 57.2 & 39.3 & 41.2 \\
8B + RL & \ding{51} & \ding{55} & 56.8 & 37.4 & 42.2 \\
8B + RL & \ding{51} & \ding{51} & \textbf{59.4} & \textbf{41.4} & \textbf{44.6} \\
\bottomrule
\end{tabular}
\caption{Within-benchmark effectiveness of Qwen3-4B-Thinking-2507 and Qwen3-8B on 402 held-out Combined Retail tasks. Values are point estimates; 5-seed mean$\pm$std for the 4B rows is reported in Appendix Table~\ref{tab:main_results_std}. Train BTM denotes structured BTM context used during RL; inference-time BTM uses across-episode online belief updating. Section~\ref{sec:btm_ablation} separately compares fixed and online variants on the full 1,339-task set.}
\label{tab:main_results}
\end{table}
\renewcommand{\arraystretch}{1}

\paragraph{Structured recovery context provides the largest zero-training gain.}
Online-updated BTM improves robustness without retraining: 20.1\%$\to$36.9\% (+16.8pp w/o alt) and 31.9\%$\to$43.6\% (+11.7pp w/ alt). Section~\ref{sec:btm_ablation} analyzes which BTM components drive this gain.

\paragraph{RL retains recovery gains without inference-time BTM.}
To isolate what the policy retains beyond inference-time recovery context, we remove BTM at inference. The RL policy reaches 26.4\%/38.8\%, versus 20.1\%/31.9\% for the base model (+6.3/+6.9pp). Base+BTM remains stronger (36.9\%/43.6\%), while combining both gives the highest performance (40.8\%/45.5\%), indicating complementary roles for learned recovery behavior and structured runtime context. Qwen3-8B shows the same qualitative pattern.

\paragraph{Full system preserves failure-free performance.}
RL+BTM clean performance (63.9$\pm$1.3\%) is within 0.4pp of the
base model (64.3$\pm$0.9\%; Appendix Table~\ref{tab:main_results_std}),
showing no measurable clean-task degradation at the resolution of this evaluation.

\paragraph{Gains persist across injection severities.}
Because our default $P(\mathrm{clean})=0.60$ setting is an intentionally severe stress test, we test whether the advantage persists under milder failure rates and different injection budgets (Table~\ref{tab:sensitivity}). As $P(\mathrm{clean})$ increases from 0.60 to 0.95, both systems improve, but RL+BTM retains a +9.0--22.2pp advantage. The gain also remains positive across stochastic injection limits (+19.9--28.3pp).

\begin{table}[t]
\centering
\small
\begin{tabular}{l c c c}
\toprule
\textbf{Setting} & \textbf{Base} & \textbf{RL+BTM} & \textbf{$\Delta$} \\
\midrule
\multicolumn{4}{l}{\emph{Vary $P(\text{clean})$; $B=5$, $K_{\max}=2$}} \\
$P(\text{clean})=0.60$ & 21.6 & 43.8 & +22.2 \\
$P(\text{clean})=0.80$ & 35.3 & 56.0 & +20.7 \\
$P(\text{clean})=0.90$ & 49.0 & 61.4 & +12.4 \\
$P(\text{clean})=0.95$ & 53.2 & 62.2 & +9.0 \\
\midrule
\multicolumn{4}{l}{\emph{Vary injection limits; $P(\text{clean})=0.60$}} \\
$B=3$, $K_{\max}=1$ & 23.9 & 52.2 & +28.3 \\
$B=5$, $K_{\max}=2$ & 21.6 & 43.8 & +22.2 \\
$B=7$, $K_{\max}=3$ & 21.9 & 41.8 & +19.9 \\
\bottomrule
\end{tabular}
\caption{
Sensitivity to injection severity on 402 held-out Combined Retail tasks
(seed 42, inject w/o alt). $B$ is the maximum number of stochastic
injections per episode; $K_{\max}$ is the maximum number of consecutive
failures on the same tool. Values are pass rates (\%).
}
\label{tab:sensitivity}
\end{table}

\paragraph{Learned recovery behavior shows small positive point-estimate gains across domains.}
Without retraining or target-domain BTM, the Retail-trained RL policy improves injected pass rate by +2.7pp on Airline-3I and +1.5pp on BFCL; adding target-domain BTM increases the gains to +9.0pp and +5.0pp, respectively. On Telecom, gains are within noise, and the 4B model is already near its clean-performance floor. This pattern is consistent with RL learning partially reusable recovery behavior while runtime context supplies domain-specific recovery information (Appendix~\ref{app:transfer}).

\subsection{BTM and RL Address Different Failure Regimes}
\label{sec:error_analysis}

\begin{table}[t]
\centering
\small
\setlength{\tabcolsep}{5pt}
\begin{tabular}{l ccc cc}
\toprule
& \multicolumn{3}{c}{\textbf{Pass Rate (\%)}} &
\multicolumn{2}{c}{\textbf{Incremental $\Delta$ (pp)}} \\
\cmidrule(lr){2-4} \cmidrule(lr){5-6}
\textbf{Error Type} & Base & +BTM & RL+BTM & $\Delta$BTM & $\Delta$RL \\
\midrule
\multicolumn{6}{l}{\emph{Service/response explicit errors}} \\
~~timeout              & 12.4 & 32.5 & 38.1 & \textbf{+20.0} & +5.6 \\
~~rate limit           & 15.2 & 33.1 & 40.5 & \textbf{+18.0} & +7.3 \\
~~server error         &  6.2 & 31.2 & 34.8 & \textbf{+25.0} & +3.6 \\
~~malformed response   & 11.5 & 36.8 & 37.8 & \textbf{+25.3} & +1.0 \\
\midrule
\multicolumn{6}{l}{\emph{Auth/schema explicit errors}} \\
~~auth error           & 13.5 & 25.9 & 36.1 & \textbf{+12.4} & +10.2 \\
~~schema drift         & 14.9 & 27.5 & 40.0 & \textbf{+12.6} & +12.5 \\
\midrule
\multicolumn{6}{l}{\emph{Silent corruptions}} \\
~~partial               & 20.6 & 29.1 & 33.6 & \textbf{+8.5} & +4.5 \\
~~stale                 & 28.9 & 31.6 & 37.0 & +2.7 & \textbf{+5.4} \\
~~factual error         & 33.8 & 29.2 & 39.8 & $-$4.6 & \textbf{+10.5} \\
\bottomrule
\end{tabular}
\caption{Pass rate by failure type under injection without alternatives.
BTM denotes the inference-time context intervention; RL+BTM is the full system.
Incremental deltas are $\Delta$BTM = $(+\mathrm{BTM})-\mathrm{Base}$ and
$\Delta$RL denotes the additional difference associated with RL training when inference-time BTM is held present. Values are point estimates. BTM uses the same online-updated inference protocol as Table~\ref{tab:main_results}.}
\label{tab:by_error}
\end{table}

The aggregate gains in Section~\ref{sec:main_results} mask substantial heterogeneity across failure types (Table~\ref{tab:by_error}). BTM provides its largest gains on timeout, rate-limit, server, and malformed-response failures (+18.0 to +25.3pp), where explicit failure signals make the supplied recovery guidance directly actionable.

The relative contribution of BTM and RL differs for the remaining error types. The RL-associated increment is +10.2/+12.5pp on auth and schema errors and +10.5pp on factual corruption, while BTM's incremental gain is smaller and becomes negative for factual corruption (-4.6pp). We treat this pattern as descriptive heterogeneity across injected error types rather than evidence of distinct underlying failure mechanisms.

\subsection{BTM Decomposition and Training Recipe Comparison}
\label{sec:btm_ablation}

\begin{table}[!htbp]
\centering
\footnotesize
\setlength{\tabcolsep}{5pt}
\renewcommand{\arraystretch}{1.05}

\begin{tabular}{
    l rrr
    @{}p{2.0em}@{}
    l cc rr
}
\toprule

\multicolumn{4}{c}{\textbf{(a) Training-recipe comparison}}
&
&
\multicolumn{5}{c}{\textbf{(b) BTM structural decomposition}}
\\[2pt]

\multicolumn{2}{c}{}
&
\multicolumn{2}{c}{\textbf{Inj (\%)}}
&
&
\multicolumn{3}{c}{}
&
\multicolumn{2}{c}{\textbf{Pass (\%)}}
\\[-2pt]

\cmidrule(lr){3-4}
\cmidrule(lr){9-10}

\textbf{Method}
& \textbf{Clean}
& \textbf{No Alt}
& \textbf{Alt}
&
&
\textbf{Variant}
& \textbf{S}
& \textbf{D}
& \textbf{Fix.}
& \textbf{Onl.}
\\
\midrule

Base, no FT
& 65.3 & 23.2 & 26.4
&
&
Base (no belief)
& \ding{55} & \ding{55} & 23.2 & -- \\

vanilla SFT
& 65.3 & 24.0 & 24.5
&
&
Dynamic only
& \ding{55} & \ding{51} & 24.9 & -- \\

vanilla GRPO
& 64.9 & 23.0 & 24.4
&
&
Static + uniform init.
& \ding{51} & \ding{55} & 33.7 & 40.8 \\

Ours (RL+BTM)
& 65.3 & \textbf{45.9} & \textbf{47.9}
&
&
Full
& \ding{51} & \ding{51} & 37.8 & \textbf{41.6} \\

\bottomrule
\end{tabular}

\caption{
(a) Training-recipe comparison on the full 1,339-task Combined Retail set. Only \emph{Ours} uses BTM (RL exploration prior + inference-time context); vanilla SFT/GRPO use no BTM at inference.
(b) BTM structural decomposition on the same 1,339-task set (inject w/o alt). 
S/D = Static recovery structure / training-derived dynamic priors; Fix./Onl.\ = dynamic posterior beliefs held fixed at their training-split estimates vs.\ updated after each completed evaluation episode using observed trajectory recovery outcomes. Both panels report point estimates with no seed-level variance. For the ``Static + uniform init.'' variant, dynamic probabilities are initialized uniformly rather than from training rollouts; in the online condition, these uniform initial values are updated from subsequent evaluation trajectories.
}
\label{tab:btm_ablation}
\end{table}

With fixed context, static recovery structure contributes most: Static + uniform init.\ improves pass rate by +10.5pp, whereas dynamic priors alone add +1.7pp. Online updating raises the former to 40.8\%, close to Full BTM at 41.6\%, showing that trajectory-based adaptation can substantially strengthen initially uninformative beliefs. Because the static condition combines generic recovery constraints with domain-specific fallback relations, their individual contributions remain unresolved.

Standard fine-tuning alone does not improve robustness: vanilla SFT and GRPO remain within 1pp of the base model under injection without alternatives (Table~\ref{tab:btm_ablation}(a)). Our full recipe reaches 45.9\%. On the same full-set condition, this is +8.1pp over Base+Full BTM with fixed beliefs and +4.3pp with online beliefs. Because this evaluation includes the 937 training tasks, we treat these differences only as a consistency check; the held-out estimate of the RL-associated gain is the +6.3pp result in Section~\ref{sec:main_results}.

\section{Discussion}
\label{sec:discussion}

\paragraph{Controlled solvability and complementary recovery mechanisms.}
Scenario control disentangles environment recoverability from policy behavior, while the results show complementary contributions from runtime recovery context and trained recovery behavior. 

\paragraph{Implications and boundaries.}

BTM should be viewed primarily as structured recovery context: static structure is most useful when context is fixed, while online updating increases the value of dynamic beliefs. The evidence is also strongest for retry and switch, given the reward design's incomplete abstention signal (Section~\ref{sec:reward}). For deployment, the main implication is practical rather than architectural: tool-failure robustness should be evaluated separately from clean task success, and jointly with efficiency, because recovery improves pass rate while increasing token use. Recovery context increases inference cost relative to the base model, but RL does not uniformly add further overhead: with alternatives, RL+BTM uses fewer tokens and steps than Base+BTM while achieving a higher pass rate (Appendix~\ref{app:btm_cost}).

\paragraph{Limitations.}
Our failures are simulated rather than incident-derived, although gains persist under milder injection settings. S3 receives no positive reward for correct stopping, so we do not establish independently learned abstention. Finally, each RL model is represented by a single training run; evaluation seeds capture rollout and injection randomness, not training-seed variance.

\section{Related Work}
\label{sec:related}

\paragraph{Tool-Using LLM Agents and Agentic RL.}
Tool-augmented LLMs have been studied via self-supervised API training~\citep{schick2023toolformer}, retrieval-augmented tool use~\citep{patil2024gorilla}, large-scale API benchmarks~\citep{qin2024toolllm,li2023apibank}, and multi-turn task completion~\citep{yao2025taubench}, with evaluation largely centered on nominal tool execution rather than controlled tool-failure recovery. On the optimization side, GRPO~\citep{shao2024deepseekmath} drops the critic via group-relative advantages, and DAPO~\citep{dapo} adds asymmetric clipping and dynamic filtering; we build on DAPO with KL stabilization and a scenario-controlled curriculum for multi-turn recovery.

\paragraph{Robustness Benchmarks for Tool-Using Agents.}
Recent work studies tool blocking~\citep{liu2026planbenchxl}, noisy environments~\citep{wang2026agentnoisebench,gupta2026reliabilitybench}, dynamic replanning~\citep{zhu2026toolsfailbenchmarkingdynamic}, and imperfect APIs~\citep{kim2026wildagteval}. ToolBench-X~\citep{tian2026toolbenchx} is closest to our injection design; \bench{} instead organizes failures by observability and controls whether recovery requires retrying, switching, or facing no viable path, enabling both controlled evaluation and training (Appendix~\ref{app:error_coverage}).

\paragraph{Training for Tool-Use Robustness.}
NoisyAgent~\citep{chen2026noisyagent}, the closest concurrent work, injects RL-rollout noise with curriculum scheduling. RobustBench-TC~\citep{zhou2026robustbench} trains via domain-randomized GRPO but targets observation-, action-, and reward-level perturbations, excluding tool-response failures from training and probing them only zero-shot with a single, 100\%-rate injection at the first call. Neither controls, during training or evaluation, whether recovery requires staying on the original path, switching, or facing no viable path; \bench{} adds this recoverability structure and explicit alternatives, letting the regimes be analyzed separately (our abstention signal remains reward-incomplete, Section~\ref{sec:reward}). PALADIN~\citep{vuddanti2026paladin} trains on failure-injected, recovery-annotated trajectories with exemplar retrieval but does not control scenario solvability; ToolCritic~\citep{hamad2025toolcritic} corrects errors via a separate critic rather than training the policy end-to-end. We instead train a single policy across episodes that each demand a different recovery regime.

\paragraph{Recovery, Abstention, and Posterior-Based Memory.}
At inference time, ReAct~\citep{yao2023react} and Reflexion~\citep{shinn2023reflexion} add retry/reflection, though LLMs cannot reliably self-correct without external feedback~\citep{huang2024large}. Abstention has been studied at the knowledge level (R-Tuning~\citep{zhang-etal-2024-r}) and the user-intent level (Learning to Ask~\citep{wang2025learningtoask}); we formulate an analogous stopping condition at the \emph{tool-response} level, subject to the same reward-design limitation (Section~\ref{sec:reward}). Bayesian-Agent~\citep{wu2026bayesianagent} maintains a posterior over reusable prompt-side skills between episodes; BTM instead supplies per-tool$\times$error posteriors as runtime context, with optional across-episode online updating, complementary rather than competing uses of a similar idea. Risk-oriented testing shows agents may fail to act on their own risk knowledge~\citep{ruan2024toolemu,tang2025agentrisk}; we study whether structured context and policy training together improve on this.

\section{Conclusion}
\label{sec:conclusion}

We presented \bench{}, a framework for studying tool-failure robustness as a recovery-policy problem rather than persistence alone. Scenario-controlled solvability separates original-path-viable, switch-required, and no-viable-path regimes, exposing a broad robustness gap across seven models. On held-out Retail tasks using the benchmark's original tool set, structured recovery context improves robustness by +16.8pp, RL retains a +6.3pp gain without inference-time context, and combining the two yields a +20.7pp improvement over the base model without measurable clean-task degradation. Overall, robust tool use appears to benefit from combining structured runtime recovery context with recovery behavior acquired through training.

\section*{AI Use Statement}

Generative AI tools were used to assist with editing and polishing the
manuscript for clarity, concision, and readability. The authors reviewed
and verified all AI-assisted edits and take full responsibility for the
final content of the paper.

\section*{Reproducibility Statement}

All experiments use publicly available benchmarks (\taubench{}, Retail-3I, Airline-3I, BFCL) and open-weight base models (Qwen3-4B-Thinking-2507, Qwen3-8B). Appendix~\ref{app:config} gives the complete training configuration; Appendix~\ref{app:injection} gives the per-tool noise distribution, injection budget, and consecutive-failure limits; Appendix~\ref{app:noise_types} defines the nine noise types; Appendix~\ref{app:alt_tools} enumerates the alternative tools; and Appendix~\ref{app:btm_prompt} gives the verbatim BTM prompt template. We will release the injection framework, domain registry, belief-computation scripts, and evaluation harness.

\section*{Ethics Statement}

This work studies failure recovery in tool-using agents using only simulated tool failures in sandboxed benchmarks; no real APIs or user data are involved. We see a primarily beneficial safety application: agents that recover from tool failures without fabricating completion once recovery paths are exhausted could reduce unwarranted actions in deployment. Because our reward does not positively reinforce correct stopping (Section~\ref{sec:reward}), this remains a design goal rather than a demonstrated property of the trained policy. Two risks merit note: training for persistence could encourage excessive retrying against real services, though our efficiency discount and repetition penalty are intended to bound this; and an escalation threshold appropriate in one domain may transfer poorly to another, since the stopping prescription is supplied by the prompt rather than learned from an abstention reward. Our Telecom results illustrate this transfer limitation: recovery training does not remedy a capability ceiling, and such systems should not be presumed robust outside their evaluated domains.

\bibliographystyle{unsrtnat}
\bibliography{references}

\newpage
\appendix

\section{Failure Injection Details}
\label{app:failure_details}

\subsection{\bench{} Injection POMDP Formalization}
\label{app:injection_algorithm}

We formalize the injected environment as a POMDP
$\mathcal{M}_{\text{inj}}=(\mathcal{S},\mathcal{A},\mathcal{O},
T_{\text{inj}},Z_{\text{inj}},R,\gamma)$ that shares the clean benchmark's
reward $R$, but whose transition and observation kernels depend on which
noise type $\nu$ is realized for a call to tool $j$.

We distinguish noise types by whether the underlying tool call is committed.
Let $\mathcal{N}_{\text{exec}}$ denote the six \emph{explicit-signal}
failures (timeout, rate limit, server error, auth error, malformed response,
schema drift) and $\mathcal{N}_{\text{silent}}$ the three \emph{silent}
corruptions (partial data, stale values, factual errors). For
$\nu \in \mathcal{N}_{\text{exec}}$, and likewise for episode-persistent
scenario blocking (Section~\ref{sec:scenarios}), the call does not execute:
the agent receives only a failure message, and the underlying state is
unaffected. For $\nu=\text{clean}$ or
$\nu \in \mathcal{N}_{\text{silent}}$, the call executes normally and only
the reported observation may be altered:

\begin{align}
T_{\text{inj}}(s_{t+1}\mid s_t,a_t,\nu) &=
\begin{cases}
T_{\text{clean}}(s_{t+1}\mid s_t,a_t)
& \nu \in \{\text{clean}\} \cup \mathcal{N}_{\text{silent}} \\
\delta(s_{t+1}=s_t)
& \nu \in \mathcal{N}_{\text{exec}} \text{ or scenario-blocked}
\end{cases} \\
Z_{\text{inj}}(o_t \mid s_t,a_t,\nu) &=
\begin{cases}
C_{\nu}\!\left(Z_{\text{clean}}(s_t,a_t)\right)
& \nu \in \{\text{clean}\} \cup \mathcal{N}_{\text{silent}} \\
C_{\nu}(\cdot)
& \nu \in \mathcal{N}_{\text{exec}} \text{ or scenario-blocked}
\end{cases}
\label{eq:pomdp_observation}
\end{align}

where $C_{\text{clean}}$ is the identity map, $\delta$ denotes a no-op
transition, and for $\nu \in \mathcal{N}_{\text{exec}}$ or scenario blocking,
$C_{\nu}(\cdot)$ is the corresponding failure or blocked-response message
rather than a corruption of an executed response.

A practical consequence is that retrying after an explicit-signal failure or
a scenario-blocked call (e.g., a \texttt{cancel\_order} call that returns a
timeout) does not risk a duplicate state mutation, whereas retrying after a
silent corruption re-executes the tool and can return an independently
corrupted value. This is a simplifying assumption rather than a model of
transactional failure semantics (e.g., a request that fails only after
partially committing side effects); we do not simulate that intermediate case.

For calls not subject to scenario blocking, each tool call to tool $j$
independently samples
$\nu_t \sim \text{Cat}(\boldsymbol{\theta}_j)$ with
$P(\text{clean})=0.60$; the remaining 0.40 is spread over the nine failure
modes in $\mathcal{N}_{\text{exec}} \cup \mathcal{N}_{\text{silent}}$.
The complete scenario-controlled injection procedure,
Algorithm~\ref{alg:bench2robust}, is given in Section~\ref{sec:scenarios}.

\subsection{Noise Types}
\label{app:noise_types}

Table~\ref{tab:noise_types} details the nine noise types used by \bench{},
organized by observability.

\begin{table}[ht]
\centering
\small
\begin{tabular}{ll}
\toprule
\textbf{Noise Type} & \textbf{Injected Response} \\
\midrule
\multicolumn{2}{l}{
\emph{Observable Errors --- explicit error signal visible to the agent}} \\
Timeout & ``Error: request timed out after 30s'' \\
Rate Limit & HTTP 429 Too Many Requests \\
Server Error & HTTP 500/502/503 Internal Server Error \\
Malformed Response & JSON parse failure / garbled output \\
Auth Error & HTTP 401/403 Forbidden \\
Schema Drift & ``Unknown field `X' / required field missing'' \\
\midrule
\multicolumn{2}{l}{
\emph{Silent Errors --- valid response with corrupted decision-critical fields}} \\
Partial & Critical fields $\to$ empty ($[\,]$, $\{\}$, \texttt{""}) \\
Stale & State fields backdated (e.g., ``delivered'' $\to$ ``pending'') \\
Factual Error & Values $\pm$10--25\% plausible perturbation \\
\bottomrule
\end{tabular}
\caption{
Nine noise types grouped by observability. Observable errors produce
unambiguous error messages; silent errors return well-formed responses with
corrupted semantics that require verification or cross-referencing to detect.
}
\label{tab:noise_types}
\end{table}

\subsection{Injection Configuration}
\label{app:injection}

Table~\ref{tab:priors} reports the per-tool noise distribution
$\boldsymbol{\theta}_j$ used throughout training and evaluation, together
with the stochastic injection budget and consecutive-failure limits
referenced in Section~\ref{sec:scenarios}.

\begin{table}[t]
\centering
\scriptsize
\setlength{\tabcolsep}{2.5pt}
\begin{tabular}{l cccccc ccc c cc}
\toprule
& \multicolumn{6}{c}{\textbf{Observable Errors}}
& \multicolumn{3}{c}{\textbf{Silent Errors}}
& & \multicolumn{2}{c}{\textbf{Limits}} \\
\cmidrule(lr){2-7}
\cmidrule(lr){8-10}
\cmidrule(lr){12-13}
& Timeout
& Rate Lim.
& Server
& Malform.
& Auth
& Schema
& Partial
& Stale
& Factual
& Clean
& Budget
& Consec. \\
\midrule
$P(\nu)$
& 0.06 & 0.05 & 0.05 & 0.04 & 0.03 & 0.03
& 0.05 & 0.04 & 0.05
& 0.60 & 5 & 2 \\
\bottomrule
\end{tabular}
\caption{
Per-tool noise configuration ($\boldsymbol{\theta}_j$). Each tool call is
clean with probability 0.60; otherwise one of nine failure modes is sampled.
Budget and consecutive-failure limits apply only to stochastic injection, not
episode-persistent scenario blocking. This severe stress-test setting is not
an estimate of real API failure rates.
}
\label{tab:priors}
\end{table}

\subsection{Alternative Tool Paths}
\label{app:alt_tools}

For Retail we add five alternative tools to the standard 16-tool environment
(21 tools total when \texttt{with\_alternatives=True}). Each provides a
different query path to equivalent task-relevant information.

\begin{table}[t]
\centering
\small
\begin{tabularx}{\columnwidth}{
@{}
>{\raggedright\arraybackslash}p{4.2cm}
>{\raggedright\arraybackslash}p{3.5cm}
>{\raggedright\arraybackslash}X
@{}
}
\toprule
\textbf{Alternative Tool} &
\textbf{Equivalent To} &
\textbf{Different Path} \\
\midrule
\texttt{search\_product\_by\_name}
& \texttt{get\_product\_details}
& Lookup by name vs.\ ID \\

\texttt{get\_item\_by\_product\_\hspace{0pt}options}
& \texttt{get\_item\_details}
& Find item by options vs.\ ID \\

\texttt{get\_orders\_for\_user}
& \texttt{get\_order\_details}
& Retrieve all orders at once (summary) \\

\texttt{get\_user\_by\_order}
& \texttt{find\_user\_id\_by\_*}
& Authenticate via order number \\

\texttt{check\_order\_item\_\hspace{0pt}availability}
& \texttt{get\_product} + matching
& One-shot exchange availability check \\
\bottomrule
\end{tabularx}
\caption{Retail alternative tools (5 added; 21 tools total).}
\label{tab:alt_tools_retail}
\end{table}

\section{BTM and Training Details}
\label{app:training_details}

\subsection{Bayesian Tool Memory Posterior and Prompt Template}
\label{app:btm_prompt}

For each (tool $j$, error type $e$) pair, recovery probability is modeled as
\[
p_{j,e}^{\text{rec}}
\sim
\text{Beta}(\alpha_{j,e}, \beta_{j,e}),
\]
with posterior mean
$\hat{P}=\alpha/(\alpha+\beta)$, computed for the Retail intervention from
base-model rollouts on the Retail training split (937 tasks, 3--5 seeds;
held-out never used). This is an \emph{episode-level recoverability}
estimate: ``when tool $j$ hits error $e$, how often does the episode still
succeed?''

Switch beliefs $\hat{P}(\text{switch}\mid j\to k)$ are defined analogously
for (source, target) pairs. These are observed recovery rates following
$j\to k$ transitions, not causal switch-success probabilities, and carry
higher uncertainty ($n=6$--$20$) due to sparser data and selection bias.

Figure~\ref{fig:btm_prompt} shows the BTM block injected into the system
prompt. Posterior recoverability statistics are presented together with
predefined fallback structure and recovery constraints; the model chooses
actions from this combined context. The template is instantiated per domain:
tool names, belief values, and fallback maps are populated from the target
environment's configuration.

\begin{figure}[t]
\centering
\fbox{\parbox{0.94\textwidth}{\ttfamily\footnotesize
<tool\_reliability> \\
Tool recovery beliefs (Beta posterior, from historical data): \\[4pt]
\{TOOL\_NAME\_1\}: \\
~~P(episode recovery | error): timeout=\{p\}, rate\_limit=\{p\}, \\
~~~~server\_error=\{p\}, auth\_error=\{p\}, schema\_drift=\{p\}, \\
~~~~malformed\_response=\{p\}, partial=\{p\}, stale=\{p\}, \\
~~~~factual\_error=\{p\} \\
~~Alternative: \{ALT\_TOOL\} (P=\{p\}, \{GRANULARITY\}) \\[2pt]
\{TOOL\_NAME\_2\}: \\
~~P(episode recovery | error): timeout=\{p\}, ... \\
~~Alternative: \{ALT\_TOOL\} (P=\{p\}, \{GRANULARITY\}) \\[2pt]
... \\[6pt]
How to use: Compare P(recovery) vs P(switch) for your error. \\
~~High P(recovery) -> retry this tool. \\
~~Low P(recovery) + high P(switch) -> switch to alternative. \\
~~After switching, verify returned data has needed fields. \\
~~If both low -> ask the user or transfer. \\[6pt]
Alternative tool paths (when primary tool fails): \\
- \{PRIMARY\} -> \{FALLBACK\} (\{GRANULARITY NOTE\}) \\
- ... \\[6pt]
Critical constraints: \\
1. If a tool fails, retry at least once before other options. \\
2. After switching, verify returned data provides ALL required \\
~~~fields at the right granularity. \\
3. Before any irreversible action, confirm every required field \\
~~~from actual tool responses. Do not guess missing fields. \\
4. For ANY price calculation, use the calculate tool. \\
5. After calling a write tool, check the response for errors. \\
6. Only transfer after exhausting retry and alternative paths. \\
</tool\_reliability>
}}
\caption{
BTM system prompt template. Placeholders are populated per domain with
recovery beliefs, fallback maps, and shared recovery constraints.
}
\label{fig:btm_prompt}
\end{figure}

\subsection{Training Configuration and Curriculum}
\label{app:config}

Table~\ref{tab:curriculum} details the five-phase scenario curriculum, and
Table~\ref{tab:config} lists the complete DAPO training configuration
referenced in Section~\ref{sec:curriculum}.

\begin{table}[t]
\centering
\small
\setlength{\tabcolsep}{3pt}
\renewcommand{\arraystretch}{1.05}
\begin{tabularx}{\columnwidth}{
@{}
c
>{\raggedright\arraybackslash}p{2.0cm}
c
c
>{\raggedright\arraybackslash}X
@{}
}
\toprule
\textbf{Phase} &
\textbf{Focus} &
\textbf{Max Inj.} &
\textbf{S1/S2/S3} &
\textbf{Advance when} \\
\midrule
1 &
S1: retry &
2 &
100/0/0 &
Avg.\ reward $>0.25$ for 3 iters \\

2 &
Discover alternatives &
2 &
80/20/0 &
Switch $>25\%$ \& pass $>25\%$ \\

3 &
S2: switch &
5 &
60/40/0 &
Switch $>25\%$ \& pass $>25\%$ \\

4 &
S3: no viable path &
5 &
40/50/10 &
S3 escalation $>50\%$, turns $<15$, no S1/S2 regression \\

5 &
Mixed &
5 &
30/45/25 &
Early stopping \\
\bottomrule
\end{tabularx}
\caption{
Scenario curriculum and phase-advancement diagnostics. Phase 4 uses the
S3-conditional transfer rate as a training diagnostic, not an abstention
correctness measure; Phase 5 early-stops on sustained over-escalation.
}
\label{tab:curriculum}
\end{table}

In the 4B training run, escalation rises from 23.7\% before S3 exposure to
33.3\% in later phases, with average turns stabilizing at 10.7. We report
this only as a curriculum diagnostic, not as an abstention-correctness
measure. Without the KL constraint, training collapses within 25--30
iterations as repetition rate exceeds 80\%.

\begin{table}[t]
\centering
\small
\begin{tabularx}{\columnwidth}{@{}l X@{}}
\toprule
\textbf{Parameter} & \textbf{Value} \\
\midrule
Base model
& Qwen3-4B-Thinking-2507 \\

Algorithm
& DAPO (group-relative with asymmetric clip + KL stabilization) \\

KL loss
& low\_var\_kl, coefficient 0.02 \\

Asymmetric clipping
& $\epsilon_{\text{low}}=0.20$, $\epsilon_{\text{high}}=0.28$ \\

Dynamic group filtering
& Groups with $\sigma < 0.01$ masked \\

Learning rate
& $2 \times 10^{-6}$ (constant) \\

Global batch size
& 256 (16 prompts $\times$ 16 rollouts) \\

Rollouts per prompt
& 16 (first 8 clean + last 8 noisy, unified normalization) \\

Max response tokens
& 8192 \\

Max steps per episode
& 25 \\

Context length
& 32768 \\

Epochs
& 3 \\

Training data
& 937 Combined Retail tasks \\

Reward
& partial\_credit $\times 0.3$ (fail) or
  $1.0 \times$ efficiency (pass) \\

Repetition penalty
& 3$\times$ same call $\to 0.5$;
  4$\times$ $\to 0.0$ \\

Efficiency discount
& $\max(0.3,\;1.0 - 0.02\max(0,\text{turns}-12))$ \\

Curriculum
& Phase 1--5 (Phases 1--4 auto-advance; Phase 5 early stopping) \\

Bayesian Tool Memory
& Beta posteriors computed from training-split rollouts \\

Hardware
& 8$\times$H100 80GB \\

User simulator
& Qwen3-235B via AWS Bedrock \\

Framework
& slime \\
\bottomrule
\end{tabularx}
\caption{Full DAPO training configuration.}
\label{tab:config}
\end{table}

\section{Error-Type Coverage of Prior Robustness Benchmarks}
\label{app:error_coverage}

Table~\ref{tab:error_coverage} situates our noise taxonomy against prior
robustness benchmarks, grouped by observability.

\begin{table}[h]
\centering
\scriptsize
\setlength{\tabcolsep}{3pt}
\begin{tabular}{l cccccc ccc}
\toprule
& \multicolumn{6}{c}{\textbf{Observable Errors}}
& \multicolumn{3}{c}{\textbf{Silent Errors}} \\
\cmidrule(lr){2-7}
\cmidrule(lr){8-10}
\textbf{Benchmark}
& Timeout
& Rate Lim.
& Server Err.
& Malformed
& Auth Err.
& Schema Dr.
& Partial
& Stale
& Factual \\
\midrule
ToolBench-X~\citep{tian2026toolbenchx}
& \ding{51} & \ding{55} & \ding{51} & \ding{55} & \ding{55}
& \ding{51} & \ding{55} & \ding{55} & \ding{51} \\

RobustBench-TC~\citep{zhou2026robustbench}
& \ding{51} & \ding{51} & \ding{51} & \ding{51} & \ding{51}
& \ding{51} & \ding{55} & \ding{55} & \ding{55} \\

AgentNoiseBench~\citep{wang2026agentnoisebench}
& \ding{51} & \ding{55} & \ding{51} & \ding{51} & \ding{55}
& \ding{55} & \ding{51} & \ding{55} & \ding{51} \\

NoisyAgent~\citep{chen2026noisyagent}
& \ding{51} & \ding{55} & \ding{51} & \ding{51} & \ding{55}
& \ding{55} & \ding{51} & \ding{55} & \ding{51} \\

ToolMaze~\citep{zhu2026toolsfailbenchmarkingdynamic}
& \ding{51} & \ding{51} & \ding{55} & \ding{55} & \ding{55}
& \ding{55} & \ding{51} & \ding{55} & \ding{51} \\

WildAGTEval~\citep{kim2026wildagteval}
& \ding{55} & \ding{55} & \ding{51} & \ding{55} & \ding{55}
& \ding{55} & \ding{51} & \ding{55} & \ding{55} \\

ReliabilityBench~\citep{gupta2026reliabilitybench}
& \ding{51} & \ding{51} & \ding{55} & \ding{51} & \ding{55}
& \ding{51} & \ding{51} & \ding{55} & \ding{55} \\

PlanBench-XL~\citep{liu2026planbenchxl}
& \ding{55} & \ding{55} & \ding{51} & \ding{55} & \ding{55}
& \ding{55} & \ding{55} & \ding{55} & \ding{55} \\

PALADIN~\citep{vuddanti2026paladin}
& \ding{51} & \ding{55} & \ding{51} & \ding{55} & \ding{55}
& \ding{55} & \ding{55} & \ding{55} & \ding{51} \\
\midrule
\bench{} (Ours)
& \ding{51} & \ding{51} & \ding{51} & \ding{51} & \ding{51}
& \ding{51} & \ding{51} & \ding{51} & \ding{51} \\
\bottomrule
\end{tabular}
\caption{
Error-type coverage across observable and silent failure modes.
\ding{51} indicates explicit coverage; \ding{55} indicates no reported
coverage.
}
\label{tab:error_coverage}
\end{table}

\section{Additional Results}
\label{app:additional_results}

\subsection{Extended Within-Benchmark Results}
\label{app:8b_results}

Table~\ref{tab:main_results} in the main text reports the Qwen3-8B
replication alongside the primary 4B target on the same 402-task Combined
Retail held-out split. Table~\ref{tab:main_results_std} gives
mean$\pm$std over five evaluation seeds for the 4B rows; the 8B rows have no
independent seed replication and are therefore omitted here.

\begin{table}[t]
\centering
\small
\begin{tabular}{l cc ccc}
\toprule
\textbf{Model}
& \textbf{Train BTM}
& \textbf{Infer BTM}
& \textbf{Clean}
& \textbf{Inj w/o Alt}
& \textbf{Inj w/ Alt} \\
\midrule
4B
& \ding{55} & \ding{55}
& 64.3{\scriptsize$\pm$0.9}
& 20.1{\scriptsize$\pm$1.4}
& 31.9{\scriptsize$\pm$2.3} \\

4B
& \ding{55} & \ding{51}
& 65.3{\scriptsize$\pm$0.9}
& 36.9{\scriptsize$\pm$1.2}
& 43.6{\scriptsize$\pm$1.6} \\

4B + RL
& \ding{51} & \ding{55}
& 64.5{\scriptsize$\pm$0.8}
& 26.4{\scriptsize$\pm$1.2}
& 38.8{\scriptsize$\pm$2.2} \\

4B + RL
& \ding{51} & \ding{51}
& 63.9{\scriptsize$\pm$1.3}
& \textbf{40.8}{\scriptsize$\pm$1.6}
& \textbf{45.5}{\scriptsize$\pm$3.0} \\
\bottomrule
\end{tabular}
\caption{
Qwen3-4B-Thinking-2507 results on the 402 held-out Combined Retail tasks
over five evaluation seeds. Clean includes evaluation-rollout randomness;
injected settings additionally include injection randomness. Corresponds to
the 4B rows of Table~\ref{tab:main_results}.
}
\label{tab:main_results_std}
\end{table}

The 8B results reproduce the qualitative complementarity between runtime BTM
and RL observed for 4B (Table~\ref{tab:main_results}). Because the 8B
checkpoint is non-thinking whereas the 4B target is a Thinking model, this
replication varies both scale and reasoning mode and should not be
interpreted as a controlled scaling comparison.

\subsection{Cross-Benchmark Transfer}
\label{app:transfer}

\begin{table}[t]
\centering
\small
\begin{tabular}{l l cc c}
\toprule
\textbf{Model}
& \textbf{Train $\to$ Eval}
& \textbf{Train BTM}
& \textbf{Infer BTM}
& \textbf{Inj w/o Alt} \\
\midrule
\multicolumn{5}{l}{\emph{BFCL multi-turn (200 tasks)}} \\
4B & --- $\to$ BFCL & \ding{55} & \ding{55} & 13.5 \\
4B & --- $\to$ BFCL & \ding{55} & \ding{51} & 16.5 \\
4B + RL & Retail $\to$ BFCL & \ding{51} & \ding{55} & 15.0 \\
4B + RL & Retail $\to$ BFCL & \ding{51} & \ding{51} & \textbf{18.5} \\
\midrule
\multicolumn{5}{l}{\emph{Airline-3I (584 tasks)}} \\
4B & --- $\to$ Airline & \ding{55} & \ding{55} & 30.7 \\
4B & --- $\to$ Airline & \ding{55} & \ding{51} & 36.6 \\
4B + RL & Retail $\to$ Airline & \ding{51} & \ding{55} & 33.4 \\
4B + RL & Retail $\to$ Airline & \ding{51} & \ding{51} & \textbf{39.7} \\
\midrule
\multicolumn{5}{l}{\emph{$\tau^2$-bench Telecom (114 tasks)}} \\
4B & --- $\to$ Telecom & \ding{55} & \ding{55} & 19.3 \\
4B & --- $\to$ Telecom & \ding{55} & \ding{51} & 21.1 \\
4B + RL & Retail $\to$ Telecom & \ding{51} & \ding{55} & 20.2 \\
4B + RL & Retail $\to$ Telecom & \ding{51} & \ding{51} & \textbf{21.9} \\
\bottomrule
\end{tabular}
\caption{
Cross-benchmark transfer for Qwen3-4B-Thinking-2507 under inject w/o alt.
RL is trained only on Retail. The (\ding{51},\ding{55}) rows evaluate the
Retail-trained policy without target-domain BTM; the
(\ding{51},\ding{51}) rows additionally supply target-domain BTM.
Values are point estimates with no seed-level variance.
}
\label{tab:transfer}
\end{table}

The Retail-trained RL policy without target-domain BTM is +2.7pp above the
base model on Airline (33.4\%) and +1.5pp on BFCL (15.0\%). Adding
target-domain BTM raises the corresponding gains to +9.0pp and +5.0pp.
On Telecom, where the 4B clean pass rate is only 21.1\%
(Table~\ref{tab:cross_benchmark}), gains are within noise.

\subsection{Inference Cost}
\label{app:btm_cost}

Table~\ref{tab:btm_cost} reports inference cost for the main Qwen3-4B
configurations on one evaluation run over the 402-task Combined Retail
held-out set (seed 42).

\begin{table}[ht]
\centering
\footnotesize
\setlength{\tabcolsep}{4pt}
\begin{tabular}{l rr rr rr}
\toprule
& \multicolumn{2}{c}{\textbf{Clean}}
& \multicolumn{2}{c}{\textbf{Inj w/o Alt}}
& \multicolumn{2}{c}{\textbf{Inj w/ Alt}} \\
\cmidrule(lr){2-3}
\cmidrule(lr){4-5}
\cmidrule(lr){6-7}
\textbf{Model}
& \textbf{Tokens} & \textbf{Steps}
& \textbf{Tokens} & \textbf{Steps}
& \textbf{Tokens} & \textbf{Steps} \\
\midrule
Base
& 88K & 11.9
& 70K & 10.3
& 67K & 9.9 \\

Base + BTM
& 103K & 12.2
& 110K & 12.8
& 123K & 12.5 \\

RL + BTM
& 105K & 12.3
& 117K & 13.4
& 108K & 11.7 \\
\bottomrule
\end{tabular}
\caption{
Aggregate token usage and mean episode length for the main Qwen3-4B
configurations on the 402-task Combined Retail held-out set (seed 42).
BTM-based configurations generally use more tokens under injection,
consistent with additional recovery actions such as retries and
alternative-tool calls.
}
\label{tab:btm_cost}
\end{table}

\section{Qualitative Examples}
\label{app:qualitative}

We present two episodes illustrating retry and switching behaviors observed
in the RL model. Each example contrasts the RL model with the base model on
the same task; because injection seeds may differ due to hash randomization,
these examples are illustrative rather than controlled paired comparisons.

\paragraph{Example 1: Retry (\texttt{task\_78}).}
\emph{Task:} User Yara Muller needs three actions---change address on order
\#W5056519, exchange an item in the same order, and cancel order
\#W5995614.

\emph{RL model (PASS, 10 tool calls):}
At step 2, \texttt{get\_user\_details} returns
\emph{HTTP 429 Too Many Requests}. The model immediately retries the same
call at step 3, which succeeds. It then retrieves all three order details,
modifies the address, exchanges the item, and cancels the third order,
completing all sub-tasks.

\emph{Base model (FAIL, 9 tool calls):}
The base model successfully completes the first two sub-tasks. When
\texttt{get\_order\_details} for the third order returns a schema validation
error at step 8 (``Required field `status' is missing''), it immediately
calls \texttt{transfer\_to\_human\_agents}, abandoning the final
cancellation.

\emph{Key insight:}
In this trajectory, the RL policy retries after an explicit failure and
succeeds, while the base model treats a later tool error as a global failure.

\paragraph{Example 2: Switch (\texttt{task\_58}).}
\emph{Task:} The user wants to exchange two items (coffee machine and
laptop) in a delivered order.

\emph{RL model (PASS, 10 tool calls):}
After authentication and \texttt{get\_user\_details} succeed, the model calls
\texttt{get\_order\_details}, which returns a schema-drift error
(``Unknown field `items\_v2'\,''). It switches to
\texttt{get\_orders\_for\_user}, a fallback that provides order summaries,
then uses \texttt{check\_order\_item\_availability} and
\texttt{get\_product\_details} to gather the remaining information and
complete the exchange.

\emph{Base model (FAIL, 2 tool calls):}
The first call \texttt{find\_user\_id\_by\_name\_zip} returns a malformed
response. The base model immediately calls
\texttt{transfer\_to\_human\_agents} without attempting authentication via
the alternative \texttt{find\_user\_id\_by\_email} path.

\emph{Key insight:}
In this trajectory, the RL policy switches to an alternative tool after a schema-drift error and recovers through an equivalent information path.

\end{document}